\documentclass[]{elsarticle}
\usepackage{booktabs} 
\usepackage{amsmath}
\usepackage{siunitx}
\usepackage{relsize}
\usepackage{amssymb}
\usepackage{algorithm}
\usepackage{booktabs}
\usepackage{verbatim} 
\usepackage{subfigure}
\usepackage{balance}
\usepackage{graphicx}
\usepackage{tikz}
\usepackage{algpseudocode}
\usepackage[bottom]{footmisc}
\usetikzlibrary{fit,positioning}
\usepackage[utf8]{inputenc}
\usepackage[english]{babel}
\usepackage{caption}
\usepackage{mathtools}
\usepackage[hyphens]{url}
\usepackage[bookmarks=false]{hyperref}
\usepackage{multirow}
\usepackage{svg}

\usepackage{xcolor}
\newcommand{\lanyu}[1]{\textcolor{black}{#1}}

\algnewcommand{\LineComment}[1]{\State  \(\triangleright\) #1 \hfill~}

\DeclareMathOperator*{\argmax}{arg\,max}
\newtheorem{myDef}{Definition}

\begin{document}
%


\begin{frontmatter}
\title{AOMD: An Analogy-aware Approach to \\Offensive Meme Detection on Social Media}

\author{Lanyu Shang} \ead{lshang@nd.edu}
\author{Yang Zhang} \ead{yzhang42@nd.edu}
\author{Yuheng Zha} \ead{yzha@nd.edu}
\author{Yingxi Chen} \ead{ychen45@nd.edu}
\author{Christina Youn} \ead{cyoun@nd.edu}
\author{Dong~Wang\corref{mycorrespondingauthor}} \ead{dwang5@nd.edu}
\address{Department of Computer Science and Engineering\\
University of Notre Dame, Notre Dame, IN, USA}

\cortext[mycorrespondingauthor]{Corresponding author}

\begin{abstract}
This paper focuses on an important problem of detecting \emph{offensive analogy meme} on online social media where the visual content and the texts/captions of the meme together make an \emph{analogy} to convey the offensive information. 
Existing offensive meme detection solutions often ignore the implicit relation between the visual and textual contents of the meme and are insufficient to identify the offensive analogy memes.
Two important challenges exist in accurately detecting the offensive analogy memes: i) it is not trivial to capture the analogy that is often implicitly conveyed by a meme; 
ii) it is also challenging to effectively align the complex analogy across different data modalities in a meme.
To address the above challenges, we develop \lanyu{a deep learning based Analogy-aware Offensive Meme Detection (AOMD) framework} to learn the implicit analogy from the multi-modal contents of the meme and effectively detect offensive analogy memes.
We evaluate AOMD on two real-world datasets from online social media. Evaluation results show that AOMD achieves significant performance gains compared to state-of-the-art baselines by detecting offensive analogy memes more accurately.
\end{abstract}

\begin{keyword}
Offensive Meme, Analogy-aware, Multi-modal Learning
\end{keyword}

\end{frontmatter}
	\section{Introduction}
\label{sec:intro}

As the popularity of social networks continues to increase, social media platforms become an attractive breeding ground for amplifying offensive activities (e.g., hate speech, cyberbullying). 
People are increasingly exposed to online offensive content in recent years.
For example, approximately 44\% of Americans were subjected to online hate and harassment in 2020, and 28\% of online social media users have experienced severe purposeful online harassment (e.g., sexual harassment, stalking, physical threats)\footnote{https://www.adl.org/online-hate-2020}. Social media platforms and researchers have been endeavoring to combat online offensive content. Many solutions have been developed to address cyber offensive behaviors. Examples include hate speech detection~\cite{van2018automatic}, cyber racism recognition~\cite{jakubowicz2017alt_right}, and online harassment identification~\cite{jhaver2018online}. 
In this paper, we study an important problem of detecting \emph{offensive analogy memes} on online social media where the visual content and the text/caption of the memes together make an \emph{analogy} to convey the offensive information.

Our problem is motivated by the prevalence of image-based content on online social media. Images often contain rich information and have been a key and attractive medium for people to create and share on social media. For example, an average of 95 million photos are uploaded to Instagram daily\footnote{https://www.wired.co.uk/article/instagram-doubles-to-half-billion-users}, and more than 40\% of tweets contain visual content\footnote{https://unionmetrics.com/blog/2017/11/include-image-video-tweets/}. In addition, social media posts with images are more likely to attract user's attention than those without (e.g., tweets with images can achieve 150\% more retweets than the tweets without images\footnote{https://blog.hubspot.com/marketing/visual-content-marketing-strategy}).
However, the widespread presence of images on social media also provides opportunities for the dissemination of offensive contents. In particular, sophisticated content creators increasingly favor the image as the carrier to propagate offensive information that is implicitly expressed with the accompanying text embedded in the image. Such kind of images can circumvent existing content censorship that focuses on the explicit indecent contents (e.g., sexual images, hateful vocabulary). \lanyu{Therefore, it is critical to effectively identify these image-driven offensive content to curb the spread of offensive information and reduce the propagation of extreme ideology on online platforms. } 

\begin{figure}[!htb]
\centering

\subfigure[][]{
    \centering
    \includegraphics[width=0.3\textwidth, height=0.22\textwidth]{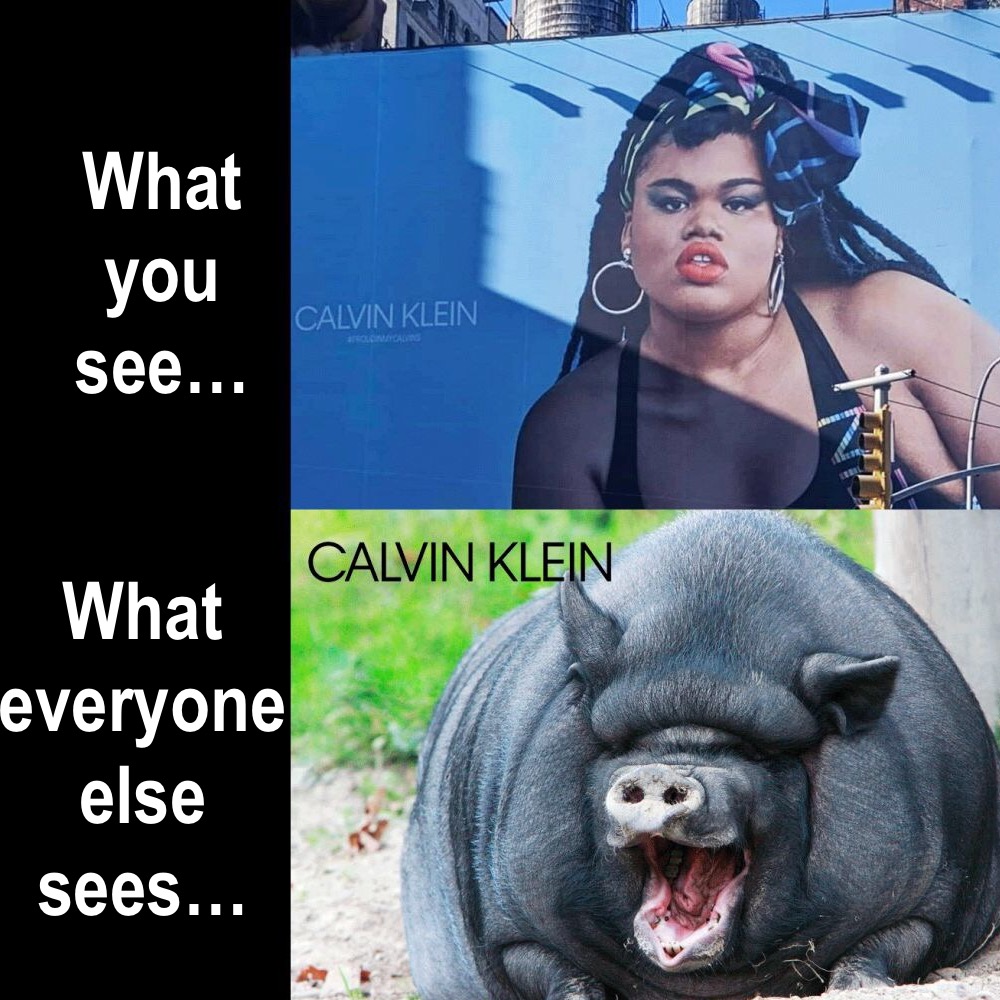}
    \label{fig:intro1}
}
\subfigure[][]{
    \centering
    \includegraphics[width=0.3\textwidth, height=0.22\textwidth]{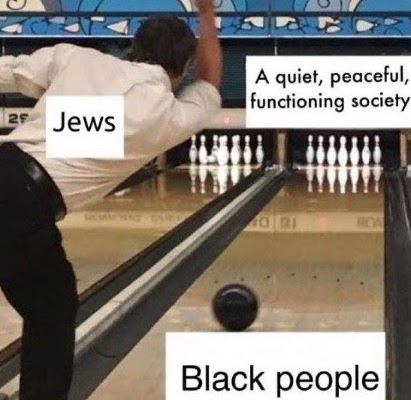}
    \label{fig:intro2}
}
\\
\subfigure[][]{
    \centering
    \includegraphics[width=0.3\textwidth, height=0.22\textwidth]{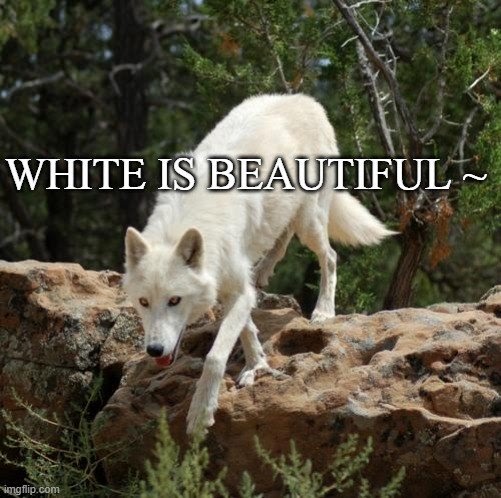}
    \label{fig:intro3}
}
\subfigure[][]{
    \centering
    \includegraphics[width=0.3\textwidth, height=0.22\textwidth]{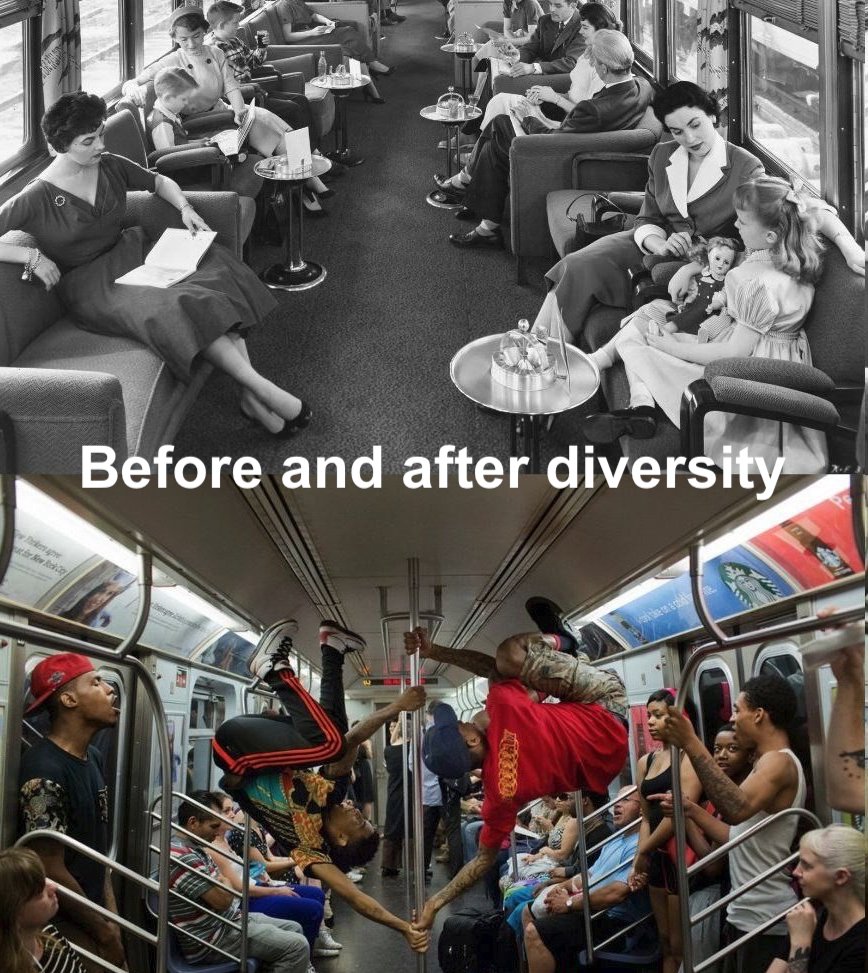}
    \label{fig:intro4}
}
\\
{\small
\begin{flushleft}
Image (a) and (b) were image-only posts. Image (c) was posted with a text description ``The world is a better place because of White man.'' Image (d) was posted with a text description ``Civilized vs. decriminalized...Ship them all to Africa.'' 
\end{flushleft}
}
\caption{Examples of Offensive Memes on Social Media}
\label{fig:intro}
\end{figure}

In this paper, we focus on an emerging phenomenon on social media where the visual content of an image together with the auxiliary text superimposed on or associated with the image jointly make an \emph{analogy} to convey offensive information to the audience of the post. We refer to such kind of content as \emph{offensive analogy meme}.
Figure \ref{fig:intro} shows four examples of the offensive analogy memes from online social media. Figure~\ref{fig:intro1} used an analogy of a black pig to taunt the appearance of a Black plus-size model. Figure~\ref{fig:intro2} delivered an implicit analogy that Jewish people are taking advantage of Black people to attack the White community. Similarly, Figure~\ref{fig:intro3} showed a favor of White privilege using an analogy of animals, and Figure~\ref{fig:intro4} leveraged the analogy of the quiet vs. energetic behavior on the train to reveal the hateful attitude against Black people. The goal of this paper is to automatically and accurately identify such offensive analogy memes on online social media.


Recently, several multi-modal solutions \cite{sabat2019hate, suryawanshi2020multimodal,sharma2020semeval,chauhan2020all,velioglu2020detecting} have been proposed to address the offensive meme detection problem on social media. However, these solutions only focus on a direct combination of the multi-modal features from the visual content and embedded captions but ignore the implicit relation between the visual and textual contents, and the analogy they deliver together.
One important observation of the above examples (Figure \ref{fig:intro}) is that these offensive analogy memes do not necessarily contain any explicit offensive or hateful content (e.g., hate speech or image) that can be leveraged to quickly detect them. Therefore, the detection of offensive analogy meme is a non-trivial problem and cannot be fully addressed by existing solutions. We elaborate the key challenges of solving this problem below.

\textit{Analogy Awareness.} 
The first challenge of detecting offensive analogy meme lies in correctly capturing and understanding the analogy expressed by the meme.  For example,  the analogy between the ``black bowling ball hits the white bowling pins'' and the ``Black people ruin the White community'' in Figure \ref{fig:intro2} is critical to detect that offensive meme. However, the extraction of such analogy often requires a holistic analysis of the visual content, embedded caption, and user comments of the meme if available~\cite{sharma2020semeval}. Moreover, the analogy of the offensive meme can also hide in the contextual information (e.g., the text description associated with the meme).
For example, Figure \ref{fig:intro3} will go undetected if we ignore the analogy between the ``WHITE'' caption in the image and the ``white man'' in the text description. Such a meme can be completely appropriate when it appears in the wildlife protection forum.
The existing solutions that focus on the image or text content itself are often insufficient to capture the analogy in such offensive memes~\cite{sharma2020semeval}.  Therefore, such analogy has to be carefully captured and considered in the process of offensive meme detection on social media.

\textit{Complex Multi-modal Analogy Alignment.}
The second challenge of detecting offensive analogy meme lies in the accurate alignment of the complex analogy across different data modalities in a meme post.
For example, existing solutions for embedded caption extraction highly rely on the optical character recognition (OCR) technique \cite{islam2017survey}. However, the OCR technique only focuses on recognizing all the characters in an image and can often recognize irrelevant content (e.g., ``CALVIN KLEIN'' in Figure \ref{fig:intro1}). Such irrelevant OCR texts may lead to the identification of incorrect analogy in the meme. Moreover, it is also important to accurately capture the analogical relation between the visual and textual content in the meme. For example, as the image shown in Figure~\ref{fig:intro2}, the implicit offensive content against Jewish people and Black people cannot be captured if the visual content and textual captions are incorrectly matched. The positions of the visual content and embedded captions have to be carefully considered to capture the analogy (i.e., bowler - ``Jews'', black bowling ball - ``Black people'', white bowling pins - ``A quiet, peaceful, functioning society'' in the above example). 
However, current multi-modal meme solutions that simply integrate visual and textual features of a meme are insufficient to capture the analogy embedded across different data modalities in the meme~\cite{kiela2020hateful}.

To address the above challenges, we develop a deep learning based Analogy-aware Offensive Meme Detection (AOMD) framework that can effectively identify offensive analogy memes on online social media. 
In particular, to address the analogy awareness challenge, we develop an analogy-aware multi-modal representation learning module to incorporate the \textit{content} (i.e., image, embedded caption) and \textit{contextual information} (i.e., text description, user comments) to identify the analogy expressed in the meme.
To address the complex multi-modal analogy alignment challenge, we develop an attentive multi-modal analogy alignment module to explicitly model the relation between the visual content and textual caption in the memes. 
To the best of our knowledge, AOMD is the first analogy-aware deep learning based solution to detect offensive analogy memes on social media. We evaluate the AOMD framework on two real-world datasets collected from Gab and Reddit. Evaluation results show that AOMD achieves significant performance gains compared to state-of-the-art baseline methods by detecting offensive analogy memes more accurately.

	\section{Related Work}

\subsection{Social Media Misbehavior}
Misbehavior has become a severe issue on social media platforms. Examples of social media misbehavior include cyberbullying~\cite{englander2017defining,van2018automatic}, trolling~\cite{paavola2016understanding}, hateful content~\cite{ribeiro2017like,magu2017detecting}, rumors~\cite{choi2020rumor}, and fake news~\cite{shu2017fake,zhang2018fauxbuster}. For example, Yao~\textit{et al.} proposed an online approach with sequential hypothesis testing to detect cyberbullying events in a timely manner~\cite{yao2019cyberbullying}. 
Cheng \textit{et al.} developed a machine learning based scheme to detect troll posts by exploring users' mood and context information on online news discussion communities~\cite{cheng2017anyone}. 
Relia \textit{et al.} developed a multi-level classifier to automatically identify targeted and self-narration of discrimination on social media~\cite{relia2019race}. 
Kumar \textit{et al.} designed a multi-task learning scheme that exploits the reply stance of social media posts to identify rumors~\cite{kumar2019tree}.
Wu \textit{et al.} developed a deep learning based framework to detect fake news by tracing the propagation pattern of posts on social media~\cite{wu2018tracing}. 
However, these solutions are insufficient to detect the offensive analogy memes on social media where the offensive content are embedded in the analogy across the visual and textual content of the meme. 
 

\subsection{Hate Speech Detection}
The spread of hate speech has gained much attention on online social media in recent years~\cite{schmidt2017survey,fortuna2018survey}. 
A number of solutions have been proposed to mitigate the problem. 
For example, Waseem \textit{et al.} proposed a critical racial theory based hate speech detection framework using n-grams and demographic information to identify racist and sexist slurs on Twitter~\cite{waseem2016hateful}. 
More recently, the phenomenon of utilizing memes (i.e., a form of multi-modal media that expresses an idea or emotion) to spread offensive content has been observed in the 2016 U.S. presidential election~\cite{suryawanshi2020multimodal}. Sabat~\textit{et~al.} developed a deep learning framework to automatically detect the hate speech in memes by fusing the visual and linguistic contents of the memes~\cite{sabat2019hate}.
\lanyu{Chauhan \textit{et al.} proposed a multi-task learning framework to simultaneously classify memes on five different tasks (e.g., offensiveness, sentiment, sarcasm)~\cite{chauhan2020all}.}
\lanyu{Velioglu \textit{et al.} developed an ensemble learning approach to boost the performance of identifying hateful memes by incorporating classification results from multiple classifiers~\cite{velioglu2020detecting}.}
However, none of the existing solutions is dedicated to study the analogy in memes, which contains critical information in identifying the offensive content. In this paper, we explicitly model the analogical relation between the visual content and embedded captions to detect offensive analogy memes on social media.


\subsection{Multi-modal Learning}
Recently, multi-modal learning has attracted much attention in learning informative features from various types of data~\cite{ramachandram2017deep}. Applications of multi-modal learning includes multi-modal machine translation \cite{specia2016shared}, visual question answering\cite{wu2017visual}, image-text matching \cite{hossain2019comprehensive}, and video description generation~\cite{jin2016video}.
For example, Zhou \textit{et al.} developed a multi-modal machine translation model that leverages visual information to assist machine translation task in distinguishing ambiguous words~\cite{zhou2018visual}. Yi \textit{et al.} proposed a deep representation learning framework to infer answers to questions of visual content~\cite{yi2018neural}. Li \textit{et al.} designed a semantic reasoning network to learn the visual representation of images and align them with text captions~\cite{li2019visual}. Hori \textit{et al.} proposed an attention-based multi-modal fusion framework to fuse image, audio, and motion features to generate video descriptions~\cite{hori2017attention}.
However, none of these multi-modal learning solutions is dedicated to identifying and aligning the analogy across different data modalities. In this work, we propose an co-attentive multi-modal analogy alignment scheme to extract and align the analogical features from the multi-modal memes to identify offensive analogy memes.

	\section{Problem Definition} \label{sec:problem}
In this section, we formally present the offensive analogy meme detection problem on social media. We first define a few key terms that will be used in the problem definition. 

\begin{figure}[!htb]
    \centering
    \includegraphics[width=0.6\textwidth]{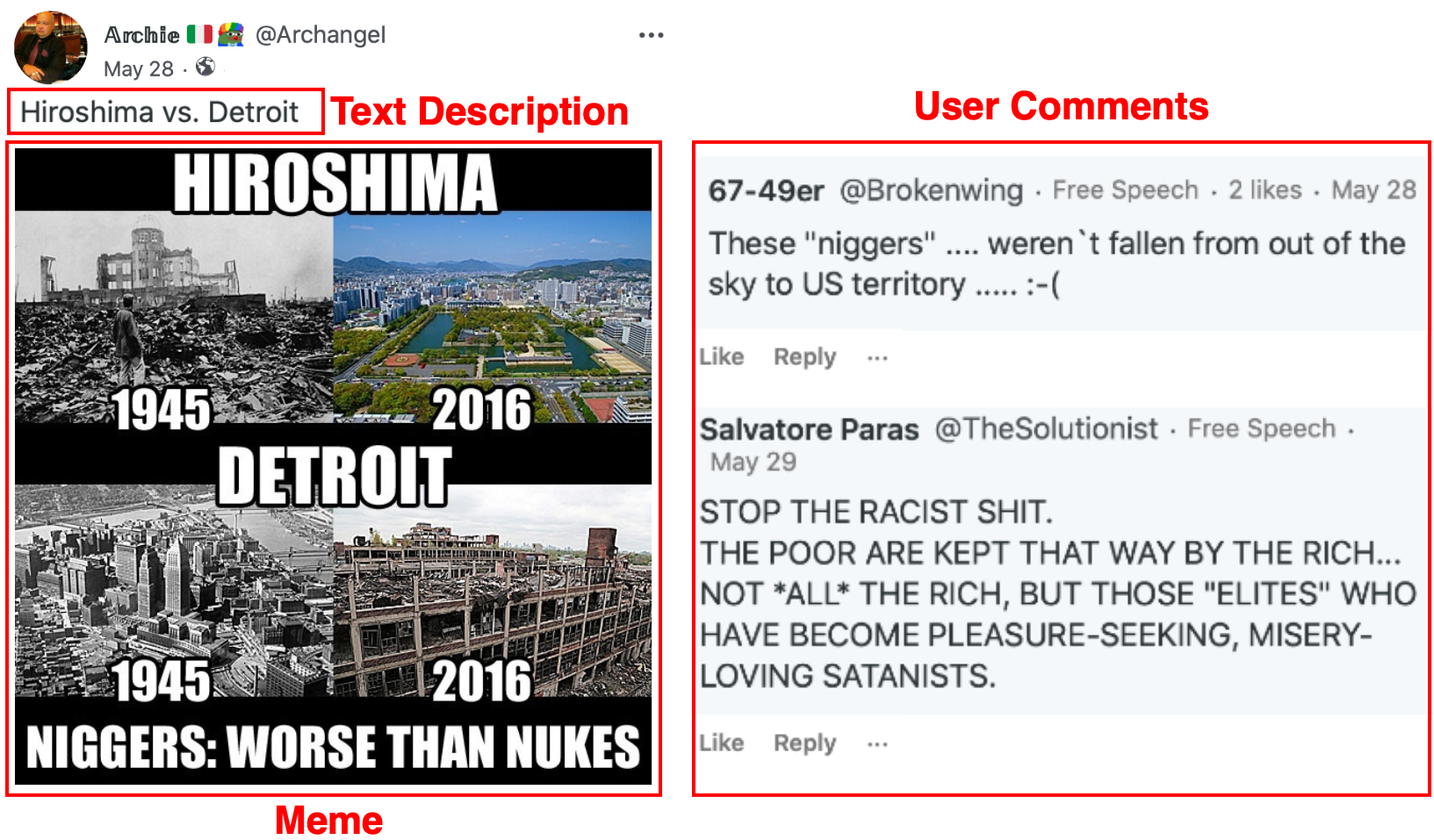}
    \caption{Example of a Meme Post on Social Media}
    \label{fig:problem} 
\end{figure}

\begin{myDef}
\emph{\textbf{Meme Post ($P_i$):} a meme post~($P_i$) on social media contains three major components: i) \textit{meme ($M_i$)}, ii) \textit{text description ($D_i$)}, and iii) \textit{user comments ($U_i$)}. An example of the meme post on social media is shown in Figure~\ref{fig:problem}.
}
\end{myDef}

\begin{myDef}
\emph{\textbf{Meme ($M_i$):} the image attachment of a meme post. The meme attachment contains two parts: the \textit{visual content ($M^V_i$)} and \textit{embedded caption ($M^C_i$)}.}
\end{myDef}

\begin{myDef}
\emph{\textbf{Visual Content ($M^V_i$)}: the imagery content in the meme that depicts visual perceptions (e.g., the view of cities in Figure \ref{fig:problem}).}
\end{myDef}

\begin{myDef}
\emph{\textbf{Embedded Caption ($M^C_i$)}: the text superimposed to or naturally contained in the meme (e.g., ``NIGGERS: WORSE THAN NUKES'' in Figure \ref{fig:problem}). }
\end{myDef}

\begin{myDef}
\emph{\textbf{Text Description ($D_i$):} the optional text description of the meme post (e.g., ``Hiroshima vs. Detroit''in Figure \ref{fig:problem}). \lanyu{The text description will be marked as an empty string for any meme post does not contain any text description.}
}
\end{myDef}

\begin{myDef}
\emph{\textbf{User Comments ($U_i$):} a set of user comments associated with the post.
}
\end{myDef}

\begin{myDef}
\emph{\textbf{Offensiveness:} the meme post is considered \textit{offensive} (denoted as $y_i = 1$) if the visual content and/or the embedded caption of the meme together with its text description (if available) conveys offensive or prejudicial information against individuals or groups of people (e.g., race, gender, religion). Otherwise, it is considered \textit{non-offensive} (denoted as $y_i = 0$).
} 
\end{myDef}


The goal of our offensive analogy meme detection is to investigate the analogy embedded in the meme and
identify its offensiveness given the meme content ($M_i$), text description ($D_i$), and user comments ($U_i$).
In particular, we assume $\mathcal{P} = \{P_1, P_2, \cdots, P_N\}$ is a set of $N$ meme posts on social media, where each meme post $P_i$ for $1\leq i \leq N$ is defined as $P_i = (M_i, D_i, U_i, y_i)$. Formally, for each $P_i \in \mathcal{P}$, our goal is to find:
\begin{equation}
    \argmax_{\hat{y_i}} Pr(\hat{y_i} = y_i | P_i), \quad \forall 1 \leq i \leq N
\end{equation}
where  $y_i$ and $\hat{y_i}$ are the ground truth and estimated label of the meme offensiveness, respectively.

%


	\section{Solution} 
\label{sec:solution}

\begin{figure*}[!htb]
    \centering
    \includegraphics[width=0.9\textwidth,height=0.4\textwidth]{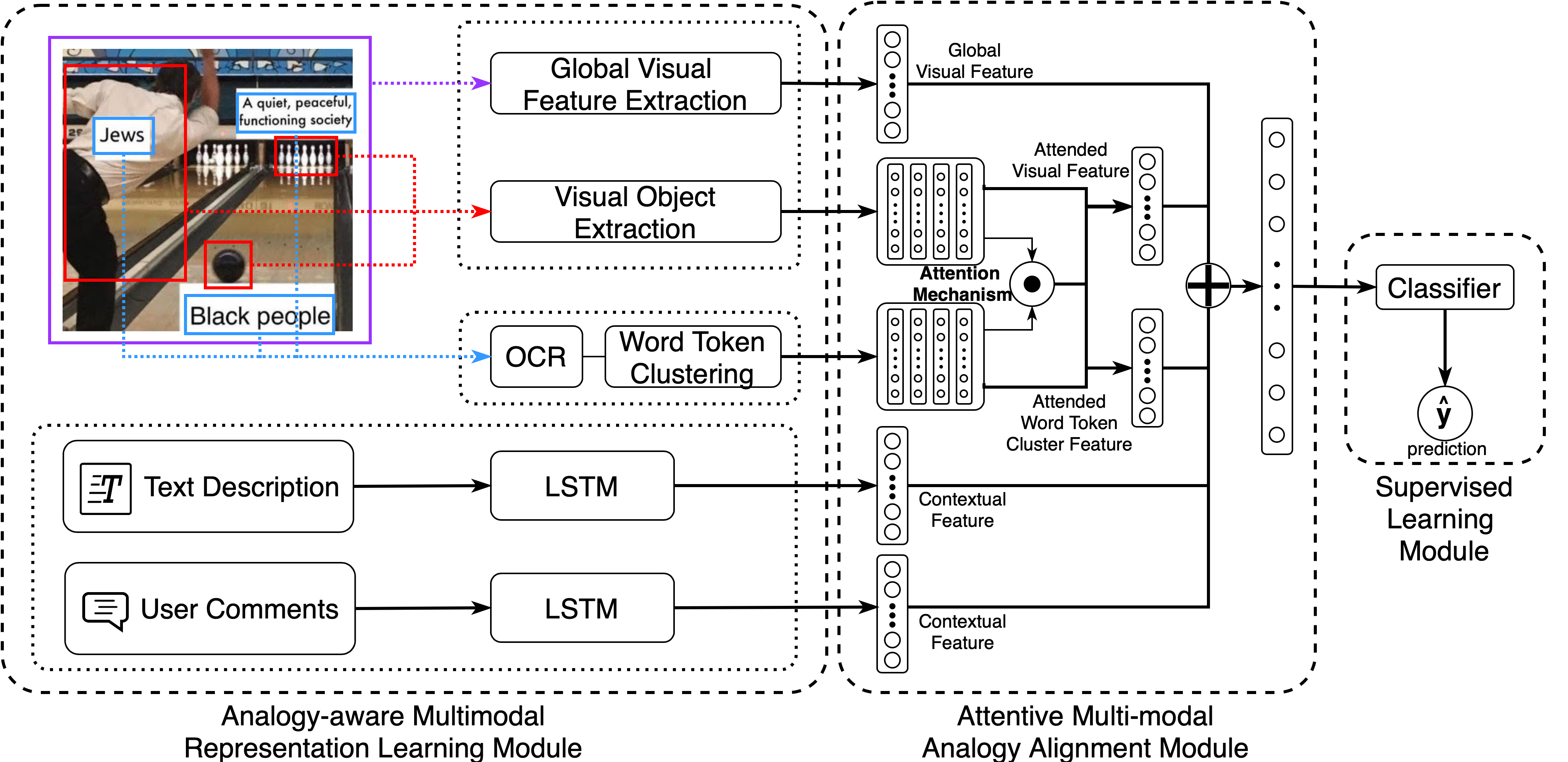}
    \caption{An Overview of the AOMD Framework}
    \label{fig:overview}
\end{figure*}

In this section, we present the Analogy-aware Offensive Meme Detection (AOMD) framework to address the offensive analogy meme detection problem defined in the previous section. An overview of the AOMD framework is shown in Figure \ref{fig:overview}. The AOMD framework contains three components. First, the \textit{analogy-aware multi-modal representation learning} module is designed to extract the visual, textual, and contextual features from the multi-modal contents of meme posts. Second, the \textit{attentive multi-modal analogy alignment} module aims to capture the analogical relation from the multi-modal content in the meme.
Third, the \textit{supervised learning} module is developed to effectively identify the offensive analogy meme in a supervised manner. We elaborate the details of each component below.

\subsection{Analogy-aware Multi-modal Representation Learning Module}
\label{sec:solution1}
The multi-modal representation learning module is designed to extract key features from each element of the meme post, and learn the representation of each element across different data modalities. In particular, we focus on the visual content, embedded caption, and contextual information (i.e., text description and user comments) to capture the offensive analogy conveyed in the meme posts.

\subsubsection{Visual Content Extraction}

The visual content in a meme image often carries essential information to identify the offensive analogy in the meme. Current solutions on offensive meme detection often focus on the image-level visual features extracted from the entire image of the meme~\cite{chen2018improving}. However, such kind of image-level features is often insufficient to capture the fine-grained visual features of the objects in the image that connect to the offensive analogy in the meme. For example, the image-level feature can capture the concept of ``person and animal'' in Figure \ref{fig:intro1}. In contrast, the object-level feature can better characterize the detailed visual feature of the ``black pig'' and ``Black model'' objects, which are the essential cues to identify the offensive analogy in that meme. In order to capture such kind of analogy embedded in the visual content, we leverage both the local object-level and global image-level information to identify offensive analogy memes.




First, we extract the local visual objects and their possible positions (i.e., the bounding box of the object) in the meme.
We observe that the visual features of objects in a meme are often relevant to the characteristics of the analog in the offensive analogy. For example, as shown in Figure \ref{fig:intro2}, the visual object ``black bowling ball'' shares the color with the embedded caption ``Black people'' and the appearance of the ``player'' (i.e., white shirt and black pants) matches the dressing style of Jewish people (i.e., the embedded caption ``Jews'' in the meme). Such visual characteristics of the local visual objects in the meme are critical to understand the offensive analogy conveyed by the meme.
In particular, the local visual objects and their positions are extracted using the advanced Faster R-CNN model~\cite{ren2015faster} pre-trained on the MSCOCO dataset~\cite{lin2014microsoft}. Formally, for each meme image, we define the set of extracted visual objects $V$ in meme $M_i$ as:
\begin{equation}
    V = \{V_1, V_2, \cdots, V_{K_i}\}
\label{eq:visual}
\end{equation}
where $V_k = (\textbf{v}_k, \textbf{p}_k)$ represents the $k^{th}$ visual object, $\textbf{v}_k\in \mathbb{R}^{d\times1}$ denotes the latent feature vector of the visual object, and $\textbf{p}_k \in \mathbb{R}^{8\times1}$ denotes the vertex coordinates of the visual object's bounding box. $K_i$ is the number of identified visual objects in the meme $M_i$.

In addition, we also extract the global image-level visual feature by encoding the image of the meme to the latent feature space with a pre-trained convolutional visual feature extractor.
We observe that the visual information of the entire image in the meme often contain valuable clues to identify the offensive analogy.
For example, the scene of ``a man bowls a ball'' in Figure \ref{fig:intro2} provides useful hints
to capture the offensive analogy of ``Jewish people leverage Black people to hit the White community''. In particular, the global visual feature is extracted from a commonly adopted deep convolutional neural network for visual recognition (i.e., ResNet50~\cite{he2016deep} pre-trained on MSCOCO). We denote the global visual feature as $\textbf{F}^{g}$.

\subsubsection{Embedded Caption Extraction}
\label{sec:sec:caption}
\begin{figure}[!htb]
\centering
\vspace{-0.1in}
\subfigure[][Offensive Meme]{
    \centering
    \includegraphics[width=0.3\textwidth, height=0.2\textwidth]{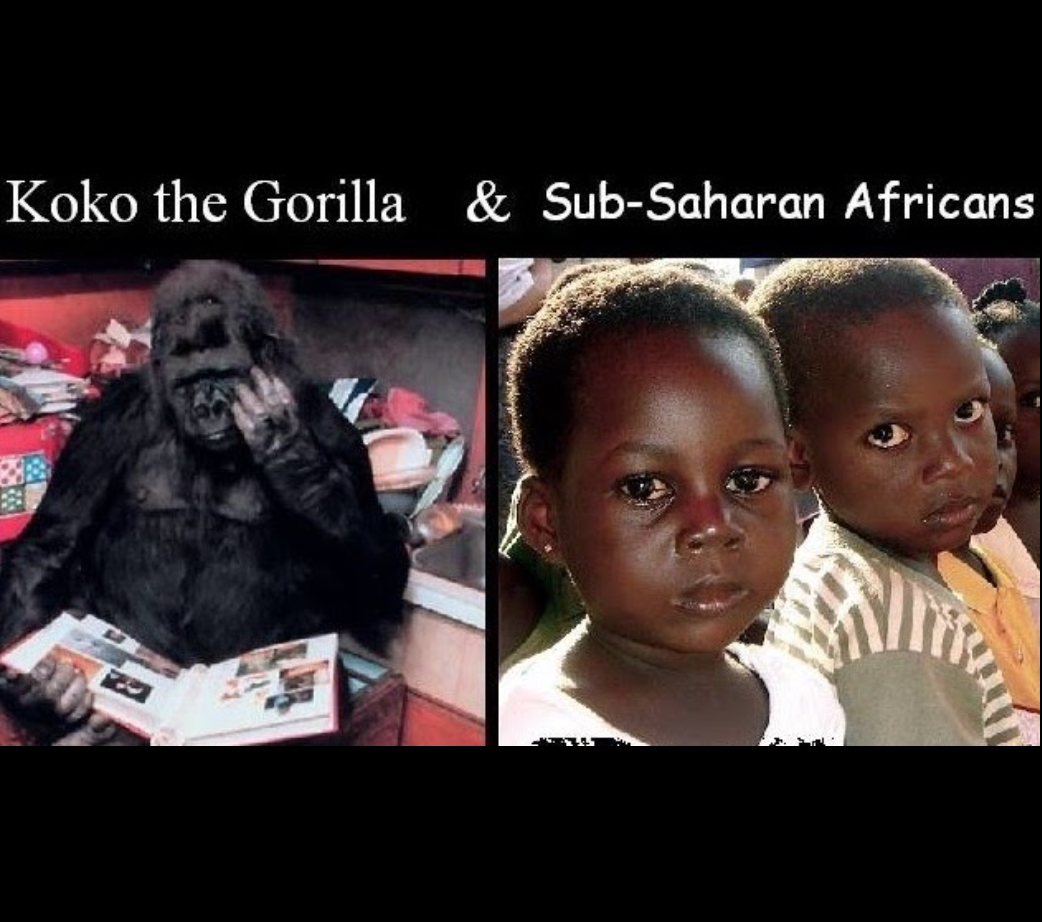}
    \label{fig:sol_example1}
}
\hspace{0.1in}
\subfigure[][Non-offensive Meme]{
    \centering
    \includegraphics[width=0.3\textwidth, height=0.2\textwidth]{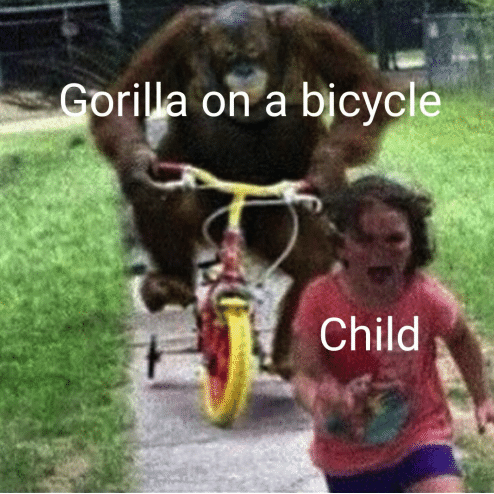}
    \label{fig:sol_example2}
}
\caption{Example of Offensive and Non-offensive Memes}
\label{fig:sol_example}
\vspace{-0.1in}
\end{figure}

In addition to the visual features, we also extract the embedded captions from the meme image.
In particular, the embedded caption is often superimposed over the image.
Different from the implicit offensive information in the visual content, captions embedded in the meme often provide explicit and indispensable information in identifying the offensive analogy. However, it is not a trivial task to identify the exact caption that contributes to an offensive analogy in a meme.  
For example, the embedded captions ``Gorilla'' and ``Africans'' in Figure \ref{fig:sol_example1} makes an analogy to deliver  the racism from the meme
while the embedded caption ``Gorilla'' and ``Child'' in Figure~\ref{fig:sol_example2} only make the meme funny.

To accurately capture the embedded caption from the meme, we first use the state-of-the-art optical character recognition (OCR) tool from Google Vision AI\footnote{https://cloud.google.com/vision} to extract the word tokens from the meme. We also record the position of the bounding box for each word token to preserve the spatial feature of the embedded captions. Formally, we denote the set of extracted word tokens $T^w$ as:
\begin{equation}
    T^w = \{T^w_1, T^w_2, \cdots\} 
\end{equation}
where $T^w_j = (w_j, \textbf{p}_j)$ represents the $j^{th}$ word token, and $w_j$ and $\textbf{p}_j \in \mathbb{R}^{8\times1}$ denote the recognized word and its bounding box coordinates, respectively.

A limitation of existing image OCR tools is that they primarily focus on the recognition of individual words in the image and are insufficient to capture the word-to-phrase association. For example, the embedded captions in Figure \ref{fig:intro2} are recognized by the OCR tool as ``A quiet, peaceful, Jews functioning society Black people'' which messes up the correct word-to-phrase association as ``Jew'', ``A quiet, peaceful, functioning society'', ``Black people''.
To overcome such a limitation, we design a cluster-based OCR to effectively extract the phrases (i.e., token cluster) from the meme. The word tokens are clustered based on the spatial position of the bounding boxes. In particular, word tokens belong to the same token cluster if their bounding boxes are overlapped or close to each other.
We denote the set of extracted word token clusters $T^c$ as:
\begin{equation}
    T^c  = \{T^c_1, T^c_2, \cdots, T^c_{S} \} 
\end{equation}
where $T^c_s = (c_s, \textbf{p}_s)$ represents the $s^{th}$ word token cluster, and $c_s$ and $\textbf{p}_s \in \mathbb{R}^{8\times1}$ denote the word sequence (i.e., phrase) in the word token cluster and the word token cluster's bounding box coordinates, respectively. $S_i$ is the number of word token clusters in the meme $M_i$.

Since each cluster contains word tokens of various lengths, we further learn the fixed-length vector representation of each cluster using a recurrent neural network (RNN) based long short term memory (LSTM) word sequence encoder~\cite{graves2005framewise} with pre-trained GloVe embeddings~\cite{pennington2014glove}. 
We use the LSTM encoder because it can learn the semantic meaning of word sequences by capturing the long-term word dependency in the sequence.
Formally, for each word token cluster $T^c_s = (c_s, \textbf{p}_s)$ in $T_c$, the encoded vector representation of the word sequence is:
\begin{equation}
    \textbf{c}_s = \text{LSTM}(c_s)
\end{equation}
The set of vector representation of word token clusters of meme $M_i$ is denoted as:
\begin{equation}
    C = \{C_1, C_2, \cdots, C_{S_i}\}
\label{eq:token}
\end{equation}
where $C_s = (\textbf{c}_s, \textbf{p}_k)$ represents the $s^{th}$ word token cluster, and $\textbf{c}_s\in \mathbb{R}^{d\times1}$ and $\textbf{p}_s \in \mathbb{R}^{8\times1}$ denote the vector representation and bounding box coordinates of the word token cluster, respectively.

\subsubsection{Contextual Information Extraction}
In addition to the visual content and embedded captions, the offensiveness of an analogy meme on social media also depends on the context of the meme post. In particular, we focus on two types of contextual information of a meme post: the text description and user comments. We observe that both the text description from the content creators and the user comments from the viewers often contain helpful information in identifying the offensive analogy.
For example, the text description of the meme in Figure \ref{fig:intro3} contains important information that can help capture the analogy between the ``white wolf'' and ``white men''. In addition, the user comments in Figure \ref{fig:problem} also provides valuable clues
in assessing the offensiveness of the analogy meme (e.g., ``STOP THE RACIST SH*T''). 
We adopt the same LSTM encoder as in caption extraction to extract the linguistic features from the text description and user comments. In particular, let $d$ and $u$ be the word sequences of the text description and user comments, respectively. The encoded vector representations for the text description and user comments are denoted as:
\begin{equation}
\begin{aligned}
    \textbf{F}^d = \text{LSTM}(d); \quad
    \textbf{F}^u = \text{LSTM}(u)
\end{aligned}
\end{equation}



\subsection{Attentive Multi-modal Analogy Alignment Module}
With the multi-modal features extracted from the visual content, embedded captions, and contextual information, we now present the attentive multi-modal analogy alignment module to extract the analogical feature from the multi-modal content of the memes.
Existing multi-modal learning solutions primarily rely on the pre-training of features on uni-modal data (e.g., object detection on image data\cite{he2016deep}, BERT on text data \cite{devlin2018bert}) to integrate cross-modal information. However, such methods often ignore the analogical relation between the visual and textual contents in meme posts and are suboptimal in detecting offensive analogy memes. For example, it will significantly reduce the possibility of catching the offensive analogy memes if we simply concatenate the visual feature of the image (e.g., ``person, bowling ball, bowling pins'' in Figure \ref{fig:intro2}) and the linguistic feature of embedded captions (e.g., ``Jews a quiet, peaceful, functioning society Black people'') but ignore the analogical relation between them (e.g., ``bowling ball'' and ``Black people'').

To address the above problem, we develop an analogy-aware attention mechanism to effectively integrate the information from different components in the meme posts. The goal of the analogy-aware attention mechanism is to learn useful features from the representation of visual objects and word token clusters that are effective in identifying the analogy of offensive memes. In particular, we observe that the visual object and word token cluster that are analogically related often appear to be in close proximity in the meme. For example, the visual object ``black gorilla'' is close to the word token cluster ``Koko the Gorilla'' in Figure \ref{fig:sol_example1} to create the analogy in that meme. To capture and preserve such a spatial proximity, we first concatenate the vector representations of the visual objects and word token clusters with their normalized bounding box positions. Specifically, let $V$ and $C$  be the set of features of the visual objects and word token clusters as defined in Equation \ref{eq:visual} and \ref{eq:token}, respectively. Then the concatenated feature vectors for each $V_k \in V$ and $C_s \in C$ are defined as:
\begin{equation}
    \Tilde{\textbf{v}}_k = [\textbf{v}_k, \textbf{p}_k]\in \mathbb{R}^{\Tilde{d}\times1} 
    \mathrm{~and~}
    \Tilde{\textbf{c}}_s = [\textbf{c}_s, \textbf{p}_s]\in \mathbb{R}^{\Tilde{d}\times1} 
\end{equation}

Next, we introduce the affinity matrix in learning the attention weights in the AOMD framework. The goal of computing the pairwise affinity is to effectively capture the pairwise analogical relation between the visual objects and word token clusters in a meme.
In particular, let $\textbf{V} = [\Tilde{\textbf{v}}_1, \Tilde{\textbf{v}}_2, \cdots, \Tilde{\textbf{v}}_K] \in \mathbb{R}^{\Tilde{d} \times K}$ and $\textbf{C} = [\Tilde{\textbf{c}}_1, \Tilde{\textbf{c}}_2, \cdots, \Tilde{\textbf{c}}_S]  \in \mathbb{R}^{\Tilde{d} \times S}$ be the feature matrices for visual object and word token cluster representations, respectively. We define the pairwise affinity matrix $\textbf{E} \in \mathbb{R}^{S \times K}$ as:
\begin{equation}
    \textbf{E} = \tanh{(\textbf{C}^\intercal \textbf{W} \textbf{V}) }
\end{equation}
where $\textbf{W} \in \mathbb{R}^{\Tilde{d} \times \Tilde{d}}$ represents the weight parameters to be learned in the neural networks. 

In addition, to effectively extract the attended features,
we adopt a common co-attention strategy~\cite{lu2016hierarchical} to compute the attention map and attention weights for visual objects and word token clusters simultaneously.
Intuitively, the visual object and word token cluster features that are more relevant to the offensive analogy in the meme will be given higher attention weights.
Formally, the attention map for visual objects  ($\textbf{M}_v \in \mathbb{R}^{\Tilde{d}\times K}$) and word token clusters ($\textbf{M}_c\in \mathbb{R}^{\Tilde{d} \times S}$) are defined as:
\begin{equation}
\begin{aligned}
    \textbf{M}_v & = \tanh(\textbf{W}_v \textbf{V} + (\textbf{W}_c \textbf{C})\textbf{E}) 
    \\
    \textbf{M}_c & = \tanh(\textbf{W}_c \textbf{C} + (\textbf{W}_v \textbf{V})\textbf{E}^\intercal) 
\end{aligned}
\end{equation}
where $\textbf{W}_v$ and $\textbf{W}_c$ are the weight parameters of the attention layer.  
The attention weights for visual objects (${\alpha}_v \in \mathbb{R}^{K}$) and word token clusters (${\alpha}_c \in \mathbb{R}^{S}$) are defined as: 
\begin{equation}
\begin{aligned}
    {\alpha}_v = \text{softmax}(\textbf{w}_v^\intercal\textbf{M}_v); \quad
    {\alpha}_c = \text{softmax}(\textbf{w}_c^\intercal\textbf{M}_c)
\end{aligned}
\end{equation}
where $\textbf{w}_v$ and $\textbf{w}_c$ are the weight parameters of the attention layer.

Using the attention weights defined above, we compute the analogically attended feature representation for the extracted visual objects and word token clusters. Formally, the attended feature vector for the visual objects ($\textbf{F}^v$) and word token clusters ($\textbf{F}^c$) are represented as follows: 
\begin{equation}
\begin{aligned}
    \textbf{F}^v = \sum_{k=1}^{K} [{\alpha}_{v}]_k \textbf{v}_k; \quad
    \textbf{F}^c = \sum_{s=1}^{S}[{\alpha}_{v}]_s \textbf{c}_s
\end{aligned}
\end{equation}

Finally, we integrate the attended feature vectors for the visual objects ($\textbf{F}^v$) and word token clusters ($\textbf{F}_c$) with the set of feature representations of the visual content ($\textbf{F}^{g}$), contextual information ($\textbf{F}^d$ and $\textbf{F}^u$) that learned from Section \ref{sec:solution1}. In particular, these feature vectors are concatenated and input to the supervised learning module discussed in the next section to detect the offensive analogy memes.

\subsection{Supervised Learning Module}
With latent features fused from the visual content, embedded captions, and contextual information as discussed in the previous sections, we now perform the binary classification to identify offensive analogy memes. In particular, we input the learned features into a two-layer feed-forward neural network (i.e., multi-layer perceptron) and a Softmax output layer that predicts the probability of a meme post $P_i$ being offensive. 
Formally, the output of the AOMD framework is defined as:
\begin{equation}\label{eq:mlp}
    \hat{y} = \text{softmax}(\text{MLP}([\textbf{F}^v_i, \textbf{F}^c_i, \textbf{F}^g_i, \textbf{F}^d_i, \textbf{F}^u_i]))
\end{equation}
where $\hat y_i$ is the estimated probability of label being $1$ (i.e., offensive).

In particular, let $y_i$ be the ground truth label (i.e., $y_i \in \{0, 1\}$ where $0$ indicates non-offensive, and $1$ indicates offensive),
our learning goal is to minimize the cross-entropy loss defined as:
\begin{equation}\label{eq:loss}
     \mathcal{L}(\theta) = \frac{1}{N}\sum_{i=1}^{N}\left(y_i \log \hat{y_i} + (1-y_i) \log(1-\hat{y_i}) \right) 
\end{equation}
where $\theta$ represents the parameters of the proposed neural network as shown in Equation \ref{eq:mlp}. We adopt the Adaptive Moment Estimation with decoupled weight decay (AdamW)~\cite{loshchilov2017decoupled} optimizer to solve our optimization problem in Equation~\ref{eq:loss}.
We summarize the AOMD in Algorithm \ref{sol:algo1}. 

\begin{figure}[!htb]
\vspace{-0.1in}
\begin{minipage}{\linewidth}
\footnotesize
    \begin{algorithm}[H]
    \caption{Summary of the AOMD Framework}
    \label{sol:algo1}
    \begin{algorithmic}[1]
    \item \textbf{input:} meme post set $\mathcal{P}$
    \item \textbf{output: $\hat y$}
    \LineComment{\textit{training phase}}
    \item \textbf{initialize: $\textbf{F}$}
    \item \textbf{for} each $P_i$ in $\mathcal{P}$ \textbf{do}
    \item \quad  \textbf{initialize:} $T^c_i$, $\textbf{F}^v_i$, $\textbf{F}^c_i$
    \item \quad extract $\textbf{F}^g_i$, $V_i$, $T^w_i$, $\textbf{F}^d_i, \textbf{F}^u_i$
    \item \quad \textbf{for} each $T^w_{i,j}$ in $T^w_i$ \textbf{do}
    \item \quad \quad assign to $T^c_{i,s} \in T^c_{i} $
    \item \quad \textbf{end for}
    \item \quad extract $C_i$ from $T^c_i$
    \item \quad $\textbf{F} \leftarrow [\textbf{F}^v_i, \textbf{F}^c_i, \textbf{F}^g_i, \textbf{F}^d_i, \textbf{F}^u_i]$ 
    \item \textbf{end for}
    \item learn $\theta$ by optimizing $\mathcal{L}(\theta)$ (Eq. \ref{eq:loss})
    \LineComment{\textit{classification phase}}
    \item \textbf{initialize: $\hat y = [~]$}
    \item \textbf{for} each $P_i \in \mathcal{P}$ \textbf{do}
    \item \quad apply neural network model (Eq. \ref{eq:mlp}) to predict $\hat y_i$ 
    \item \quad $\hat y \leftarrow \hat y_i$
    \item \textbf{end for}
    \item output $\hat y$
    \end{algorithmic}
    \end{algorithm}
\end{minipage}
\end{figure}

	\section{Data}
\label{sec:data}
In this section, we present the real-world datasets and labels we collected for evaluation. We observe that mainstream social media platforms (e.g., Twitter) contain a massive amount of memes. However, the collection of offensive memes on those platforms has experienced a long tail issue (i.e., only a small portion of the collected memes are actually offensive)~\cite{zhang2019hate}. 
\lanyu{In addition, we observe that existing datasets for offensive memes (e.g., MultiOFF~\cite{suryawanshi2020multimodal}, SemEval-2020~\cite{sharma2020semeval}) only contain the meme images but lack of the text description and user comments which are essential for the detection of offensive memes. Therefore, we collected our own datasets for the comprehensive evaluation of the proposed AOMD scheme.}

We choose two offensive meme appealing social media forums, the ``Memes, memes, and more Memes'' group on Gab\footnote{https://gab.com/} and the \textit{r/NewOffensiveMemes} subforum on Reddit\footnote{https://reddit.com/}, as our data sources to collect the offensive analogy memes for our study. For each meme post, we collect the meme content, text description, and user comments. \lanyu{We note that meme images are diversified and rarely duplicated on these two forum-driven platforms where users often prefer posting ``new'' content to re-posting.}
Next, we invite three independent annotators to annotate the label of each meme (i.e., the offensiveness of the meme) by carefully assessing the analogy embedded in the meme content. The ground-truth labels are decided based on the majority of the three annotators. The inter-annotator agreement (i.e., Fleiss Kappa score \cite{warrens2010inequalities}) of the Gab and Reddit datasets are 0.47 and 0.51, respectively. 
A summary of the collected datasets is presented in Table \ref{tab:data}. While our datasets are collected from offensive meme appealing forums, we observe that our datasets are not balanced (i.e.,  non-offensive memes are more than offensive ones), which is consistent with the observation on mainstream social media platforms. We also observe that 63.2\% and 99.5\% of the meme posts in the Gab and Reddit datasets contain contextual information (i.e., text description, user comments), respectively.


\begin{table}[htb!]
\caption{\lanyu{Dataset Statistics}}
\label{tab:data}
\centering
\resizebox{0.9\columnwidth}{!}{
    \begin{tabular}{l c c }
    \toprule
    \midrule
    \textbf{Data Trace} & \textbf{Gab} & \textbf{Reddit}\\
    \toprule
    Total Number of Collected Meme Posts & 1,965 & 1,094  \\ 
    \cmidrule(){1-3}
    Number of Offensive Meme Posts & 672 (34.2\%) &  380 (34.7\%) \\ 
    \cmidrule(){1-3}
    Number of Meme Posts Containing Analogy & 891 (45.3\%) & 522 (47.7\%) \\ 
    \cmidrule(){1-3}
    Number of Meme Posts Containing Contextual Information &1,242 (63.2\%)  & 1,089 (99.5\%) \\ 
    \midrule
    \toprule
    
    \end{tabular}
}
\vspace{-0.2in}
\end{table}

\vspace{-0.2in}




	\section{Evaluation} 
\label{sec:eval}
In this section, we evaluate the performance of the AOMD framework on the real-world datasets described in Section \ref{sec:data}. In particular, we compare the performance of AOMD with the state-of-the-art baselines from the literature. The results show that the AOMD framework achieves significant performance gains in terms of the offensive analogy meme detection accuracy compared to all baselines.

\subsection{Baselines and Experiment Setting}
We compare AOMD with several state-of-the-art baselines in detecting offensive analogy memes on social media.

\begin{itemize}
    \item \textbf{WSCNet \cite{yang2018weakly}:} a convolutional neural network based scheme that learns the sentiment representation of visual content and classifies the sentiment of an image. We use the meme image as input and train the WSCNet scheme to detect offensive analogy meme.
    
    \item \textbf{HateSpeech \cite{davidson2017automated}:} a lexicon-based approach that automatically detects hate speech on social media with a set of linguistic and sentiment features. We take the embedded caption of each meme post as the input to the HateSpeech framework. 
    
    \item \textbf{MultiOFF \cite{suryawanshi2020multimodal}:} a recent transfer learning based approach that detects offensive analogy memes by fusing pre-trained visual and linguistic features into a dense neural network (i.e., MLP). 
    
    \item \textbf{HatePixel \cite{sabat2019hate}:} a multi-layer perceptron classification approach that detects hate speech in images by encoding visual and textual contents into latent vectors with pre-trained models (i.e., VGG-16 for visual representations and BERT for text embeddings). 
    
    \item \textbf{HatefulMeme \cite{kiela2020hateful}:} a recent multi-modal hateful meme detection framework that performs the detection task by leveraging the pre-trained feature from Visual BERT~\cite{li2019visualbert} with a MLP classifier.
    
\end{itemize}

We use the meme images and OCR word tokens as the inputs to the WSCNet and HateSpeech baseline, respectively. The inputs to the multi-modal baselines (i.e., MultiOFF, HatePixel, and HatefulMeme) are the same as AOMD but exclude the contextual information since none of these baselines explicitly model the contextual information.
\lanyu{Therefore, we present the performance of three ablations of the AOMD framework for the fair comparison with baseline methods taking less input than AOMD. In the meantime, we also evaluate the contribution of the attentive multi-modal analogy alignment component in the ablation study. In particular, we consider the following ablations of the AOMD framework.}

\begin{itemize}
    \item \lanyu{\textbf{AOMD w/o Visual:} the AOMD framework excludes the global visual feature extracted from the visual content of each meme.}
    \item \lanyu{\textbf{AOMD w/o OCR:} the AOMD framework excludes the OCR extraction of embedded captions from each meme.}
    \item \lanyu{\textbf{AOMD w/o Context:} the AOMD framework excludes the contextual information (i.e., text description and user comments of each meme post).}
    \item \lanyu{\textbf{AOMD w/o Attention:} the AOMD framework removes the attention layer and combines the the average visual object feature and word token cluster features by concatenation.}
\end{itemize}

For all compared baselines, we use 70\%, 10\%, and  20\% of the dataset as the training, validation, and testing set, respectively.
The vector lengths of the visual features, encoded embedded caption, text description and user comments are set to 100. We use AdamW~\cite{loshchilov2017decoupled} as the optimizer and set learning rate = 0.001, $\epsilon$=1e-8, and the batch size to 32. For all other baselines, we follow the network architectures presented in the papar and carefully tune the hyperparameters to achieve the best performance of each baseline.

\subsection{Detection Effectiveness}
In the first set of experiments, we evaluate the detection accuracy of the AOMD framework and all compared baselines. 
In particular, we use a set of common metrics for binary classification to evaluate the detection performance: \textit{Accuracy}, \textit{F1 Score}, and \textit{Cohan's Kappa Coefficient (Kappa)}. \lanyu{The results are shown in Table~\ref{tab:classification_gab} and Table~\ref{tab:classification_reddit}.}

We observe that the AOMD scheme consistently outperforms all baseline methods on all evaluation metrics on both the Gab and Reddit datasets. 
For example, AOMD outperforms the best performing baselines on the Gab dataset (i.e., MultiOFF) and Reddit dataset (i.e., HatefulMeme) by 14.9\% and 8.3\% in terms of F1 score, respectively.
Such significant performance gains of AOMD can be attributed to the accurate identification of the analogy features extracted from the multi-modal contents and the effective alignment of those features across different modalities in the meme.
We observe that the visual content based solution (i.e., WSCNet) is not robust in detecting offensive analogy memes because it ignores the text embedded in the meme and is insufficient to capture the offensive information jointly expressed by the embedded caption and visual content. Similarly, HateSpeech only focuses on the textual content in the post but ignores the semantic meanings of visual objects. Therefore, it is also suboptimal in accurately identifying the offensive analogy memes. In contrast, the AOMD framework incorporates the visual content, embedded captions, and contextual information of a meme post to capture the offensive analogy jointly conveyed by the meme. 

Furthermore, in comparison to the multimodel baseline methods (i.e., MultiOFF, HatePixel, HatefulMeme) that focus on fusing the pre-trained visual and linguistic features, we observe that AOMD and the ablated AOMD (i.e., AOMD w/o context that takes the same input as these multimodal baselines) also achieve significant performance gains.
This is because the AOMD framework is not only designed to fusing features extracted from different data modalities but also capturing their implicit analogical relations with the attention mechanism. 

\begin{figure}[!htb]
\centering
    \centering
    \includegraphics[width=0.8\textwidth]{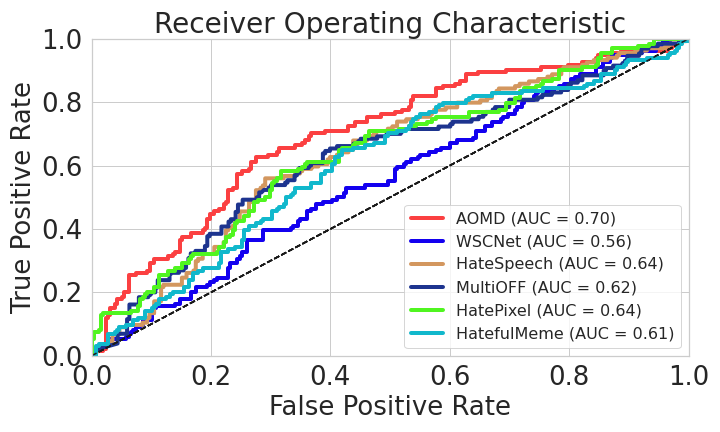}
\caption{ROC Curves of All Methods - Gab}
\label{fig:roc_gab}
\end{figure}

\begin{figure}[!htb]
\centering
    \centering
    \includegraphics[width=0.8\textwidth]{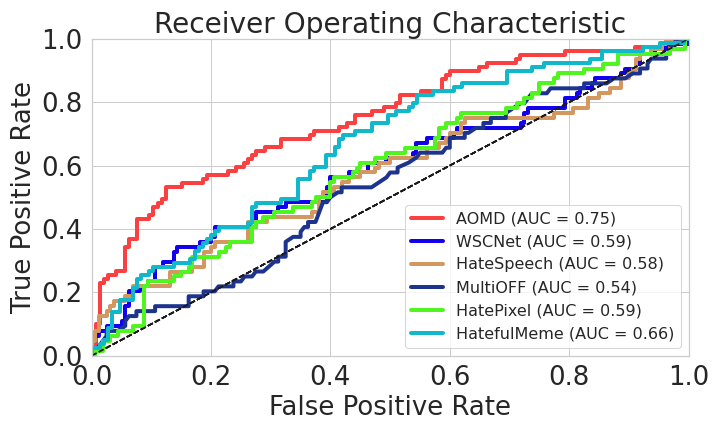}
\vspace{-0.2in}
\caption{ROC Curves of All Methods - Reddit}
\label{fig:roc_reddit}
\vspace{-0.2in}
\end{figure}

\lanyu{Finally, we plot the Receiver Operating Characteristic (ROC) curves (Figure \ref{fig:roc_gab} and Figure \ref{fig:roc_reddit}) of all compared methods to evaluate the detection performance with respect to all classification thresholds. The AOMD framework continues to outperforms all baseline methods.}

\begin{table}[!htb]
    \footnotesize
    \centering
    \caption{Classification Performance for All Methods - Gab}
    \label{tab:classification_gab}
    \resizebox{0.8\columnwidth}{!}{
    \begin{tabular}{l l  c  c  c }
    \toprule
    \midrule
    &Method &Accuracy &F1 &Kappa  \\
    \toprule
    &WSCNet &0.6045 &0.4237 &0.1228\\
    \cmidrule(r){2-5}
    &HateSpeech &0.6323 &0.4790 &0.2067\\
    \cmidrule(r){2-5}
    Baselines &MultiOFF &0.6254 &0.4844 &0.2255\\
    \cmidrule(r){2-5}
    &HatePixel &0.6377 &0.4817 &0.2035\\
    \cmidrule(r){2-5}
    &HatefulMeme &0.6173 &0.4361 &0.1465\\
    \cmidrule(){1-5}
    \multirow{2}{1.5cm}{Ablations of AOMD}
    &AOMD w/o Visual &0.6782  &0.5196 &0.2814\\
    \cmidrule(r){2-5}
    &AOMD w/o OCR &0.6454  &0.4832 &0.2476\\
    \cmidrule(r){2-5}
    &AOMD w/o Context &0.6811  &0.5318 &0.2893\\
    \cmidrule(r){2-5}
    &AOMD w/o Attention &0.6607  &0.4981 &0.2637\\
    \cmidrule(r){1-5}
    &\textbf{AOMD} &\textbf{0.6913} &\textbf{0.5568} &\textbf{0.3201}\\
    \midrule
    \toprule
    \end{tabular}
    }
\end{table}

\begin{table}[!htb]
    \footnotesize
    \centering
    \caption{Classification Performance for All Methods - Reddit}
    \label{tab:classification_reddit}
    \resizebox{0.8\columnwidth}{!}{
    \begin{tabular}{l l  c  c  c }
    \toprule
    \midrule
    &Method &Accuracy &F1 &Kappa\\
    \toprule
    &WSCNet  &0.6523 &0.3862 &0.1529\\
    \cmidrule(r){2-5}
    &HateSpeech  &0.6039 &0.3026 &0.0633\\
    \cmidrule(r){2-5}
    Baselines &MultiOFF  &0.5826 &0.2851 &0.0316\\
    \cmidrule(r){2-5}
    &HatePixel  &0.6347 &0.3659 &0.1184\\
    \cmidrule(r){2-5}
    &HatefulMeme  &0.6718 &0.4492 &0.2335\\
    \cmidrule(){1-5}
    \multirow{2}{1.5cm}{Ablations of AOMD}
    &AOMD w/o Visual &0.6915  &0.4650 &0.2662\\
    \cmidrule(r){2-5}
    &AOMD w/o OCR  &0.6734 &0.4538 &0.2397\\
    \cmidrule(r){2-5}
    &AOMD w/o Context &0.6908 &0.4627 &0.2633\\
    \cmidrule(r){2-5}
    &AOMD w/o Attention &0.6852  &0.4583 &0.2571\\
    \cmidrule(r){1-5}
    &\textbf{AOMD} &\textbf{0.7172} &\textbf{0.4867} &\textbf{0.2834}\\
    \midrule
    \toprule
    \end{tabular}
    }
\end{table}


\subsection{\lanyu{Analogy Awareness}}
\begin{figure}[!htb]
\centering
\subfigure[][Gab]{
    \centering
    \includegraphics[width=0.45\textwidth]{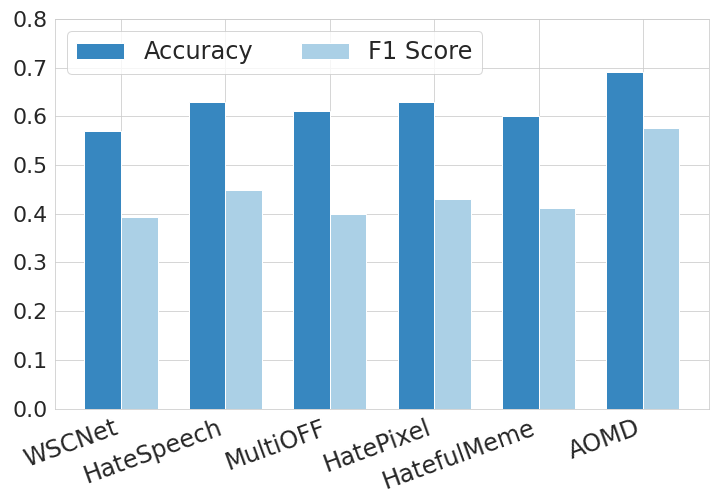}
    \label{fig:eval_analogy_gab}
}
\hspace{0.1in}
\subfigure[][Reddit]{
    \centering
    \includegraphics[width=0.45\textwidth]{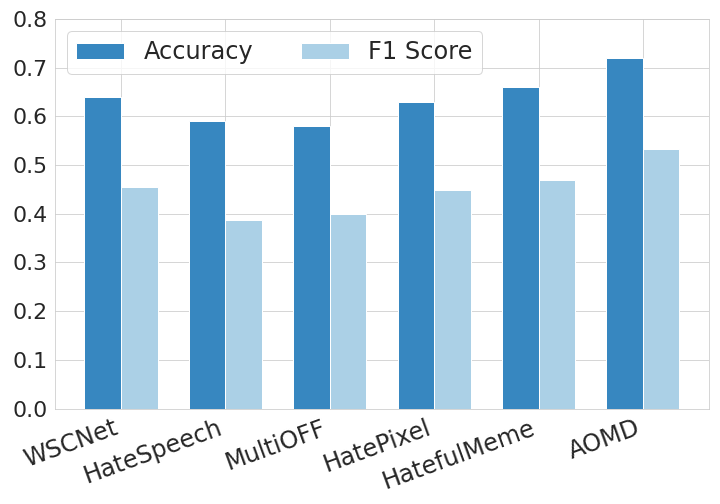}
    \label{fig:eval_analogy_reddit}
}
\caption{Detection Performance on Analogy Meme Posts}
\label{fig:eval_analogy}
\end{figure}

\lanyu{In the second set of experiments, we further investigate the effectiveness of identifying offensive meme posts containing analogy. In particular, we randomly pick 100 analogy meme posts (i.e., meme posts contains analogy) from each test set of the Gab and Reddit datasets to evaluate the detection performance. The accuracy and F1 score are summarized in Figure~\ref{fig:eval_analogy}. We observe that the performance of AOMD outperforms all the baseline schemes on both the Gab and Reddit datasets. Such performance gains again demonstrate the effectiveness of the AOMD framework in identifying offensive analogy memes by modeling the analogical relation across different data modalities in the meme posts.
}

	\section{Conclusion}

In this paper, we develop AOMD, the first analogy-aware deep learning based solution to address offensive analogy memes on social media. The AOMD framework is designed to effectively capture the analogy conveyed by different data modalities of the meme, and detect the offensiveness implicitly expressed in the meme. We evaluate the AOMD framework with two real-world datasets collected from Gab and Reddit. The evaluation results demonstrate that our scheme outperforms the state-of-the-art baselines by accurately identifying the offensive analogy memes on social media. 




\bibliographystyle{elsarticle-num}
\bibliography{refs}

\begin{thebibliography}{10}
\expandafter\ifx\csname url\endcsname\relax
  \def\url#1{\texttt{#1}}\fi
\expandafter\ifx\csname urlprefix\endcsname\relax\def\urlprefix{URL }\fi
\expandafter\ifx\csname href\endcsname\relax
  \def\href#1#2{#2} \def\path#1{#1}\fi

\bibitem{van2018automatic}
C.~Van~Hee, G.~Jacobs, C.~Emmery, B.~Desmet, E.~Lefever, B.~Verhoeven,
  G.~De~Pauw, W.~Daelemans, V.~Hoste, Automatic detection of cyberbullying in
  social media text, PloS one 13~(10) (2018) e0203794.

\bibitem{jakubowicz2017alt_right}
A.~Jakubowicz, et~al., Alt\_right white lite: trolling, hate speech and cyber
  racism on social media, Cosmopolitan Civil Societies: An Interdisciplinary
  Journal 9~(3) (2017) 41.

\bibitem{jhaver2018online}
S.~Jhaver, S.~Ghoshal, A.~Bruckman, E.~Gilbert, Online harassment and content
  moderation: The case of blocklists, ACM Transactions on Computer-Human
  Interaction (TOCHI) 25~(2) (2018) 1--33.

\bibitem{sabat2019hate}
B.~O. Sabat, C.~C. Ferrer, X.~Giro-i Nieto, Hate speech in pixels: Detection of
  offensive memes towards automatic moderation, arXiv preprint
  arXiv:1910.02334.

\bibitem{suryawanshi2020multimodal}
S.~Suryawanshi, B.~R. Chakravarthi, M.~Arcan, P.~Buitelaar, Multimodal meme
  dataset (multioff) for identifying offensive content in image and text, in:
  Proceedings of the Second Workshop on Trolling, Aggression and Cyberbullying,
  2020, pp. 32--41.

\bibitem{sharma2020semeval}
C.~Sharma, D.~Bhageria, W.~Scott, S.~PYKL, A.~Das, T.~Chakraborty,
  V.~Pulabaigari, B.~Gamback, Semeval-2020 task 8: Memotion analysis--the
  visuo-lingual metaphor!, arXiv preprint arXiv:2008.03781.

\bibitem{chauhan2020all}
D.~S. Chauhan, S.~Dhanush, A.~Ekbal, P.~Bhattacharyya, All-in-one: A deep
  attentive multi-task learning framework for humour, sarcasm, offensive,
  motivation, and sentiment on memes, in: Proceedings of the 1st Conference of
  the Asia-Pacific Chapter of the Association for Computational Linguistics and
  the 10th International Joint Conference on Natural Language Processing, 2020,
  pp. 281--290.

\bibitem{velioglu2020detecting}
R.~Velioglu, J.~Rose, Detecting hate speech in memes using multimodal deep
  learning approaches: Prize-winning solution to hateful memes challenge, arXiv
  preprint arXiv:2012.12975.

\bibitem{islam2017survey}
N.~Islam, Z.~Islam, N.~Noor, A survey on optical character recognition system,
  arXiv preprint arXiv:1710.05703.

\bibitem{kiela2020hateful}
D.~Kiela, H.~Firooz, A.~Mohan, V.~Goswami, A.~Singh, P.~Ringshia,
  D.~Testuggine, The hateful memes challenge: Detecting hate speech in
  multimodal memes, arXiv preprint arXiv:2005.04790.

\bibitem{englander2017defining}
E.~Englander, E.~Donnerstein, R.~Kowalski, C.~A. Lin, K.~Parti, Defining
  cyberbullying, Pediatrics 140~(Supplement 2) (2017) S148--S151.

\bibitem{paavola2016understanding}
J.~Paavola, T.~Helo, H.~Jalonen, M.~Sartonen, A.~Huhtinen, Understanding the
  trolling phenomenon: The automated detection of bots and cyborgs in the
  social media, Journal of Information Warfare 15~(4) (2016) 100--111.

\bibitem{ribeiro2017like}
M.~H. Ribeiro, P.~H. Calais, Y.~A. Santos, V.~A. Almeida, W.~Meira~Jr, " like
  sheep among wolves": Characterizing hateful users on twitter, arXiv preprint
  arXiv:1801.00317.

\bibitem{magu2017detecting}
R.~Magu, K.~Joshi, J.~Luo, Detecting the hate code on social media, arXiv
  preprint arXiv:1703.05443.

\bibitem{choi2020rumor}
D.~Choi, S.~Chun, H.~Oh, J.~Han, et~al., Rumor propagation is amplified by echo
  chambers in social media, Scientific Reports 10~(1) (2020) 1--10.

\bibitem{shu2017fake}
K.~Shu, A.~Sliva, S.~Wang, J.~Tang, H.~Liu, Fake news detection on social
  media: A data mining perspective, ACM SIGKDD explorations newsletter 19~(1)
  (2017) 22--36.

\bibitem{zhang2018fauxbuster}
D.~Y. Zhang, L.~Shang, B.~Geng, S.~Lai, K.~Li, H.~Zhu, M.~T. Amin, D.~Wang,
  Fauxbuster: A content-free fauxtography detector using social media comments,
  in: 2018 IEEE International Conference on Big Data (Big Data), IEEE, 2018,
  pp. 891--900.

\bibitem{yao2019cyberbullying}
M.~Yao, C.~Chelmis, D.~S. Zois, Cyberbullying ends here: Towards robust
  detection of cyberbullying in social media, in: The World Wide Web
  Conference, 2019, pp. 3427--3433.

\bibitem{cheng2017anyone}
J.~Cheng, M.~Bernstein, C.~Danescu-Niculescu-Mizil, J.~Leskovec, Anyone can
  become a troll: Causes of trolling behavior in online discussions, in:
  Proceedings of the 2017 ACM conference on computer supported cooperative work
  and social computing, 2017, pp. 1217--1230.

\bibitem{relia2019race}
K.~Relia, Z.~Li, S.~H. Cook, R.~Chunara, Race, ethnicity and national
  origin-based discrimination in social media and hate crimes across 100 us
  cities, in: Proceedings of the International AAAI Conference on Web and
  Social Media, Vol.~13, 2019, pp. 417--427.

\bibitem{kumar2019tree}
S.~Kumar, K.~M. Carley, Tree lstms with convolution units to predict stance and
  rumor veracity in social media conversations, in: Proceedings of the 57th
  Annual Meeting of the Association for Computational Linguistics, 2019, pp.
  5047--5058.

\bibitem{wu2018tracing}
L.~Wu, H.~Liu, Tracing fake-news footprints: Characterizing social media
  messages by how they propagate, in: Proceedings of the eleventh ACM
  international conference on Web Search and Data Mining, 2018, pp. 637--645.

\bibitem{schmidt2017survey}
A.~Schmidt, M.~Wiegand, A survey on hate speech detection using natural
  language processing, in: Proceedings of the Fifth International workshop on
  natural language processing for social media, 2017, pp. 1--10.

\bibitem{fortuna2018survey}
P.~Fortuna, S.~Nunes, A survey on automatic detection of hate speech in text,
  ACM Computing Surveys (CSUR) 51~(4) (2018) 1--30.

\bibitem{waseem2016hateful}
Z.~Waseem, D.~Hovy, Hateful symbols or hateful people? predictive features for
  hate speech detection on twitter, in: Proceedings of the NAACL student
  research workshop, 2016, pp. 88--93.

\bibitem{ramachandram2017deep}
D.~Ramachandram, G.~W. Taylor, Deep multimodal learning: A survey on recent
  advances and trends, IEEE Signal Processing Magazine 34~(6) (2017) 96--108.

\bibitem{specia2016shared}
L.~Specia, S.~Frank, K.~Sima’an, D.~Elliott, A shared task on multimodal
  machine translation and crosslingual image description, in: Proceedings of
  the First Conference on Machine Translation: Volume 2, Shared Task Papers,
  2016, pp. 543--553.

\bibitem{wu2017visual}
Q.~Wu, D.~Teney, P.~Wang, C.~Shen, A.~Dick, A.~van~den Hengel, Visual question
  answering: A survey of methods and datasets, Computer Vision and Image
  Understanding 163 (2017) 21--40.

\bibitem{hossain2019comprehensive}
M.~Z. Hossain, F.~Sohel, M.~F. Shiratuddin, H.~Laga, A comprehensive survey of
  deep learning for image captioning, ACM Computing Surveys (CSUR) 51~(6)
  (2019) 1--36.

\bibitem{jin2016video}
Q.~Jin, J.~Liang, Video description generation using audio and visual cues, in:
  Proceedings of the 2016 ACM on International Conference on Multimedia
  Retrieval, 2016, pp. 239--242.

\bibitem{zhou2018visual}
M.~Zhou, R.~Cheng, Y.~J. Lee, Z.~Yu, A visual attention grounding neural model
  for multimodal machine translation, arXiv preprint arXiv:1808.08266.

\bibitem{yi2018neural}
K.~Yi, J.~Wu, C.~Gan, A.~Torralba, P.~Kohli, J.~Tenenbaum, Neural-symbolic vqa:
  Disentangling reasoning from vision and language understanding, in: Advances
  in neural information processing systems, 2018, pp. 1031--1042.

\bibitem{li2019visual}
K.~Li, Y.~Zhang, K.~Li, Y.~Li, Y.~Fu, Visual semantic reasoning for image-text
  matching, in: Proceedings of the IEEE International Conference on Computer
  Vision, 2019, pp. 4654--4662.

\bibitem{hori2017attention}
C.~Hori, T.~Hori, T.-Y. Lee, Z.~Zhang, B.~Harsham, J.~R. Hershey, T.~K. Marks,
  K.~Sumi, Attention-based multimodal fusion for video description, in:
  Proceedings of the IEEE international conference on computer vision, 2017,
  pp. 4193--4202.

\bibitem{chen2018improving}
D.~Chen, H.~Li, X.~Liu, Y.~Shen, J.~Shao, Z.~Yuan, X.~Wang, Improving deep
  visual representation for person re-identification by global and local
  image-language association, in: Proceedings of the European Conference on
  Computer Vision (ECCV), 2018, pp. 54--70.

\bibitem{ren2015faster}
S.~Ren, K.~He, R.~Girshick, J.~Sun, Faster r-cnn: Towards real-time object
  detection with region proposal networks, in: Advances in neural information
  processing systems, 2015, pp. 91--99.

\bibitem{lin2014microsoft}
T.-Y. Lin, M.~Maire, S.~Belongie, J.~Hays, P.~Perona, D.~Ramanan,
  P.~Doll{\'a}r, C.~L. Zitnick, Microsoft coco: Common objects in context, in:
  European conference on computer vision, Springer, 2014, pp. 740--755.

\bibitem{he2016deep}
K.~He, X.~Zhang, S.~Ren, J.~Sun, Deep residual learning for image recognition,
  in: Proceedings of the IEEE conference on computer vision and pattern
  recognition, 2016, pp. 770--778.

\bibitem{graves2005framewise}
A.~Graves, J.~Schmidhuber, Framewise phoneme classification with bidirectional
  lstm and other neural network architectures, Neural networks 18~(5-6) (2005)
  602--610.

\bibitem{pennington2014glove}
J.~Pennington, R.~Socher, C.~D. Manning, Glove: Global vectors for word
  representation, in: Proceedings of the 2014 conference on empirical methods
  in natural language processing (EMNLP), 2014, pp. 1532--1543.

\bibitem{devlin2018bert}
J.~Devlin, M.-W. Chang, K.~Lee, K.~Toutanova, Bert: Pre-training of deep
  bidirectional transformers for language understanding, arXiv preprint
  arXiv:1810.04805.

\bibitem{lu2016hierarchical}
J.~Lu, J.~Yang, D.~Batra, D.~Parikh, Hierarchical question-image co-attention
  for visual question answering, in: Advances in neural information processing
  systems, 2016, pp. 289--297.

\bibitem{loshchilov2017decoupled}
I.~Loshchilov, F.~Hutter, Decoupled weight decay regularization, arXiv preprint
  arXiv:1711.05101.

\bibitem{zhang2019hate}
Z.~Zhang, L.~Luo, Hate speech detection: A solved problem? the challenging case
  of long tail on twitter, Semantic Web 10~(5) (2019) 925--945.

\bibitem{warrens2010inequalities}
M.~J. Warrens, Inequalities between multi-rater kappas, Advances in data
  analysis and classification 4~(4) (2010) 271--286.

\bibitem{yang2018weakly}
J.~Yang, D.~She, Y.-K. Lai, P.~L. Rosin, M.-H. Yang, Weakly supervised coupled
  networks for visual sentiment analysis, in: Proceedings of the IEEE
  conference on computer vision and pattern recognition, 2018, pp. 7584--7592.

\bibitem{davidson2017automated}
T.~Davidson, D.~Warmsley, M.~Macy, I.~Weber, Automated hate speech detection
  and the problem of offensive language, in: Eleventh international aaai
  conference on web and social media, 2017.

\bibitem{li2019visualbert}
L.~H. Li, M.~Yatskar, D.~Yin, C.-J. Hsieh, K.-W. Chang, Visualbert: A simple
  and performant baseline for vision and language, arXiv preprint
  arXiv:1908.03557.

\end{thebibliography}

\end{document}